\def\year{2018}\relax
\documentclass[letterpaper]{article} 
\usepackage{aaai18}  
\usepackage{times}  
\usepackage{helvet}  
\usepackage{courier}  
\usepackage{url}  
\usepackage{graphicx}  
\usepackage{multirow}
\usepackage{lipsum}
\usepackage{graphicx,dblfloatfix}%
\begin{document}
%

\title{Capturing Ambiguity in Crowdsourcing Frame Disambiguation}
\author{
 Anca Dumitrache \\
 Vrije Universiteit Amsterdam \\
 IBM CAS Benelux \\
 \texttt{anca.dmtrch@gmail.com} \\
 \And
 Lora Aroyo\\
 Vrije Universiteit Amsterdam \\
 IBM CAS Benelux \\
 \texttt{lmaroyo@gmail.com} \\ 
 \And
 Chris Welty\\
 Google Research, New York \\
 \texttt{cawelty@gmail.com} \\ 
}

\maketitle
\begin{abstract}
FrameNet is a computational linguistics resource composed of semantic frames, high-level concepts that represent the meanings of words. In this paper, we present an approach to gather frame disambiguation annotations in sentences using a crowdsourcing approach with multiple workers per sentence to capture inter-annotator \emph{disagreement}. We perform an experiment over a set of 433 sentences annotated with frames from the FrameNet corpus, and show that the aggregated crowd annotations achieve an F1 score greater than 0.67 as compared to expert linguists. We highlight cases where the crowd annotation was correct even though the expert is in disagreement, arguing for the need to have multiple annotators per sentence.  Most importantly, we examine cases in which crowd workers could not agree, and demonstrate that these cases exhibit ambiguity, either in the sentence, frame, or the task itself, and argue that collapsing such cases to a single, discrete truth value (i.e. correct or incorrect) is inappropriate, creating arbitrary targets for machine learning.
\end{abstract}

\section{Introduction}

FrameNet is a computational linguistics resource based on the frame semantics theory~\cite{baker1998berkeley}. A semantic {\it frame} is an abstract representation of a word sense, describing a type of entity, relation, or event, and identifies the associated \emph{roles} implied by the frame. The FrameNet resource offers a collection of semantic frames, together with a corpus of documents annotated with these frames. In the corpus, individual words are mapped to the single frame that represents the meaning of that word in the sentence.  

Since many words have multiple possible meanings, the task of obtaining these annotations is called \emph{frame disambiguation}, similarly to word-sense disambiguation.  It is a complex task that typically is performed by linguistic experts, subjected to strict annotation guidelines and quality control~\cite{baker2012framenet}. As such, this task typically does not scale sufficiently in order to meet the annotation requirements of modern machine learning methods. Moreover, the annotation is typically performed by only one expert, which makes it impossible to capture any diversity of perspectives.  

There have been a number of attempts at using crowdsourcing for frame disambiguation in sentences, such as those by \citeauthor{Hong:2011:GCR:2018966.2018970}~\shortcite{Hong:2011:GCR:2018966.2018970} and \citeauthor{chang2015scaling}~\shortcite{chang2015scaling}, offering a creative way to deal with the complexity of the annotation task. This paper addresses the considerable problem of \emph{ambiguity}  in frame annotation, which we show to be a prominent feature in frame semantics.  We adapt the CrowdTruth framework, which encourages using multiple crowd annotators to perform the same work, and processes the disagreement between them to signal low quality workers, sentences, and frames.

This paper presents the following contributions:

\begin{enumerate}

\item \emph{annotated corpus}: 433 FrameNet sentences with crowd annotations;

\item \emph{crowd vs. expert evaluation}: the crowd achieves comparative quality with trained FrameNet experts (F1 $>0.67$), and we provide examples of typical cases where the crowd annotation is correct despite the expert disagreement;

\item \emph{metrics for frame and sentence quality}: a qualitative evaluation showing that inter-annotator disagreement is an indicator of ambiguity in both frames and sentences.

\item \emph{ambiguity-aware annotation methodology}: we demonstrate that the cases in which the crowd workers could not agree exhibit ambiguity, either in the sentence, frame, or the task itself; we argue that collapsing such cases to a single, discrete truth value (i.e. correct or incorrect) is inappropriate, creating arbitrary targets for machine learning.

\end{enumerate}

\section{Related Work}
This work relates to the state of the art in two areas of research: (1) various crowdsourcing approaches for FrameNet related tasks, and (2) dealing with ambiguity and disagreement in crowdsourcing. Below we provide an overview of the research on which we base or inspire our approach.

\begin{figure*}[t!]
\centering
\caption{Fragment of the crowdsourcing task template.}
\label{fig:template}
\includegraphics[width=0.65\textwidth]{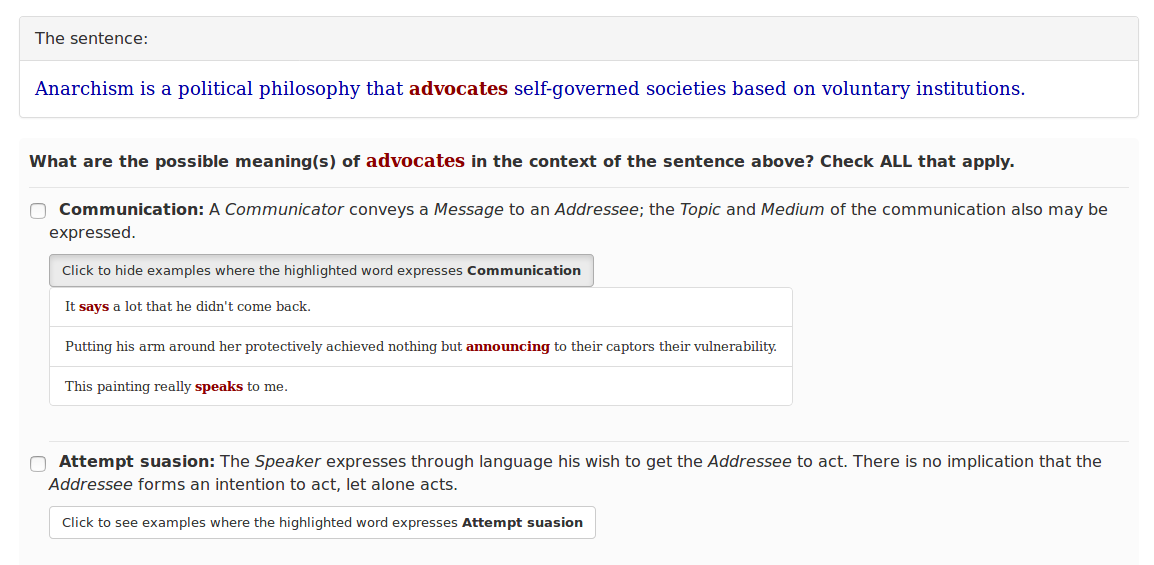}
\end{figure*}

\subsection{Crowdsourcing FrameNet}

\citeauthor{Hong:2011:GCR:2018966.2018970}~\shortcite{Hong:2011:GCR:2018966.2018970} first experimented with applying crowdsourcing for frame disambiguation, where the authors were able to achieve an accuracy of 0.982 as compared to the expert annotators. We replicate the performance of the crowd from this research in our experiments. Moreover, we also measure the inter-annotator disagreement which we show is a useful indicator of ambiguity in both sentences and frames. \citeauthor{fossati2013outsourcing}~\shortcite{fossati2013outsourcing} extend the frame disambiguation task with identifying frame roles (roles are the elements of the semantic frame, e.g. participants in an event).

More recently, \citeauthor{chang2015scaling}~\shortcite{chang2015scaling} proposed a method for supervised crowdsourcing of frame disambiguation, where after an initial step of picking the best frame for a word in a sentence, the crowd worker receives feedback from the other annotators, and can then decide if they want to change their annotation or not. This serves to correct misunderstandings of the frame definition by the crowd. \citeauthor{pavlick2015framenet+}~\shortcite{pavlick2015framenet+} use automatic paraphrasing to increase the lexical coverage of FrameNet, where crowdsourcing is employed to manually filter out bad paraphrases.

Similarly to our claim, \citeauthor{jurgens2013embracing}~\shortcite{jurgens2013embracing} argues that ambiguity is an inherent feature of frame/word sense disambiguation, and that crowdsourcing can be used to capture it. The crowd is asked to annotate on a Likert scale the degree to which a sense applies to a word. As Likert scales have been shown to be unreliable for capturing subjective measures~\cite{Kittur:2008:CUS:1357054.1357127}, our annotation task is composed of quantifiable binary questions (i.e. does the frame apply to the word in the sentence or not?), and the ambiguity is captured by giving the same examples to multiple workers and measuring disagreement~\cite{aroyo2014threesides}.

In our experiments we found between 10-15 workers provided the most reliable results (the more complex the task, the more workers are needed). Thus, we employ 15 annotators per task in our experiments in order to ensure we capture sufficient diversity of  interpretations, compared to 10 by \citeauthor{Hong:2011:GCR:2018966.2018970}~\shortcite{Hong:2011:GCR:2018966.2018970} and 3 by \citeauthor{jurgens2013embracing}~\shortcite{jurgens2013embracing}.

\subsection{Disagreement and Ambiguity in Crowdsourcing}

Our work is part of a continuous effort in exploring the inter-annotator disagreement as an indicator for (1) inherent uncertainty in the domain knowledge as \citeauthor{cheatham2014conference}~\shortcite{cheatham2014conference} found when assessing the Ontology Alignment Evaluation Initiative (OAEI) benchmark, (2) debatable cases in linguistic theory, rather than faulty annotation, as \citeauthor{plank-hovy-sogaard:2014:P14-2}~\shortcite{plank-hovy-sogaard:2014:P14-2} found in their part-of-speech tagging task, and (3) ambiguity inherent in natural language~\cite{Bayerl2011}.

In our own work, we have primarily been interested in ambiguity at the sentence level and in the target semantics~\cite{DBLP:journals/corr/DumitracheAW17}.  The CrowdTruth project has made software available~\cite{inel2014crowdtruth} to process vector representations of crowd gathered data that \emph{encourages disagreement}, in a more continuous representation of truth. We replicated our  approach from other semantic interpretations tasks to the frame disambiguation task.

Finally we note recent efforts to consider in ground truth corpora (1) the notion of uncertainty, where \citeauthor{schaekermann2016}~\shortcite{schaekermann2016} also use disagreement in crowdsourcing for modeling it, (2) the notion of ambiguity, where \citeauthor{Chang:2017:Revolt}~\shortcite{Chang:2017:Revolt} found that ambiguous cases cannot simply be resolved by better annotation guidelines or through worker quality control, and (3) the notion of noise, where \citeauthor{lin2014re}~\shortcite{lin2014re} show that machine learning classifiers can often achieve a higher accuracy when trained with noisy crowdsourcing data.

\section{Crowdsourcing Setup}

\subsection{Dataset}
The dataset used in this experiment consists of sentence-word pairs from the FrameNet corpus from release 1.7 (the latest one at the time of writing), where  the given word in the sentence has been labeled with a frame by expert annotators. We selected a word in each sentence and constructed a list of candidate frames to show to the crowd (Fig.~\ref{fig:template}). To do this, we used the Framester corpus~\cite{gangemi2016framester}, which maps FrameNet semantic frames to synonym sets from WordNet~\cite{miller1995wordnet}. First, the sentences were processed with tokenization, sentence splitting, lemmatization and part-of-speech tagging. Then each word with a frame attached to it was matched with all of its possible synonym sets from WordNet, while making sure that the part-of-speech constraint of the synonym set is fulfilled. Using the WordNet mapping, we constructed a list of possible frames for each word with an expert annotation. From this dataset, we randomly selected 433 sentence-word pairs, containing 341 unique frames and 300 unique words after lemmatization, that respect the following conditions:

\begin{itemize}

\item The word has a part-of-speech of either a {\it noun} or a {\it verb}.

\item Each word has {\it at least two and no more than 20 candidate frames}.

\end{itemize}

The restriction on the maximum number of frames was done so as not to overwhelm the crowd with too many choices. However, annotating words that have more than 20 frames can easily be adapted for our template, by fragmenting the candidate frame list into several parts and running the task multiple times. Also, having just one frame per word means that the crowdsourcing task becomes one of validation, not disambiguation, so the restriction on the minimum number of frames was put in place.

For simplicity, we refer to the sentence-word pairs as sentences in the rest of the paper. This dataset, as well as the crowdsourcing results and aggregated metrics are available online\footnote{\url{https://github.com/CrowdTruth/FrameDisambiguation}}.

\subsection{Task Template}

The crowdsourcing task was run on the Amazon Mechanical Turk platform\footnote{\url{https://mturk.com/}}. The task template is shown in Figure~\ref{fig:template}. The workers were given a sentence with the word highlighted, and then asked to perform the multiple choice task of selecting all frames that fit the sense of the highlighted word, or that none of the frames fit. The most challenging part of the frame disambiguation task design is making sure that the crowd can understand the meaning of the frame. For each frame, we show the definition, as well as a list of sentences exemplifying the usage of the frame. These example sentences can be accessed by the workers by clicking a button next to each frame, so that the workers do not become overwhelmed with the information on the task page. In order to make sure we capture diverse worker opinions, we increased the number of annotators per sentence from 10 (the number recommended by \citeauthor{Hong:2011:GCR:2018966.2018970}~\shortcite{Hong:2011:GCR:2018966.2018970}), to 15. The cost of the task varied from \$0.08 per annotation at the start of the task, in order to attract a sizable pool of workers, to \$0.06 at the end, as workers became quicker at solving the task.

\subsection{Disagreement Metrics}

To aggregate the results of the crowd, while also capturing inter-annotator disagreement, we use a modified version of the CrowdTruth~\cite{aroyo2014threesides} metrics. The first step is to construct the {\it worker vectors}, which are a set of binary vectors encoding the decision of one worker for one sentence. The vector has $n+1$ components, where $n$ is the number of frames shown together with the sentence. If the worker selects a frame from the multiple-choice list, its corresponding component would be marked with `1', and `0' otherwise. The decision to pick none of the frames also corresponds to a component in the vector. Using these worker vectors, we then calculate the following disagreement metrics:

\begin{itemize}
\item {\bf frame-sentence score (FSS):} the degree with which a frame matches the sense of the word in the sentence. It is the ratio of workers that picked the frame to all the workers that read the sentence, weighted by the worker quality (WQS).  A higher FSS should indicate that the frame is more clearly expressed in a sentence.

\item {\bf sentence quality (SQS):} the overall worker agreement over one sentence. It is the average cosine similarity over all worker vectors for one sentence, weighted by the worker quality (WQS) and frame quality (FQS). A higher SQS should indicate a clear sentence.

\item {\bf frame quality (FQS):} the agreement on a frame in all sentences that it appears. Given frame $f$, $ FQS(f) = avg(FSS(f,s) | FSS(f,s) > 0)$. $FQS$ is also weighed by the quality of the workers and the sentences.  A higher FQS should indicate a clear frame semantics.

\item {\bf worker quality (WQS):} the overall agreement of one crowd worker with the other workers, calculated using average cosine similarity with other workers per sentence, and weighted by the sentence and frame qualities.

\end{itemize}

\begin{figure*}[tbh!]
\centering
\begin{minipage}{.471\textwidth}
\centering
\caption{F1 score of crowd annotations using expert annotation as true positives}
\label{fig:sent_f1}
\includegraphics[width=\textwidth]{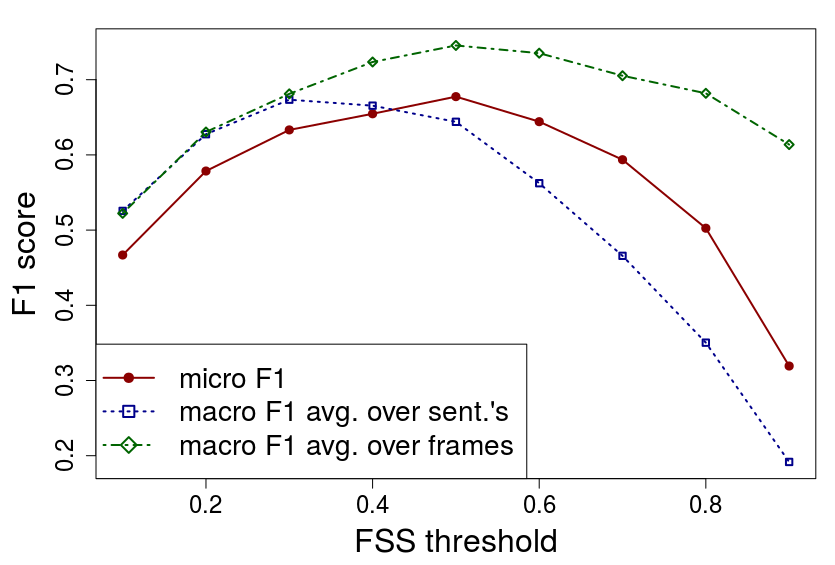}
\end{minipage} \qquad 
\begin{minipage}{.471\textwidth}
\centering
\caption{Accuracy of crowd annotations using expert annotation as correct}
\label{fig:sent_acc}
\includegraphics[width=\textwidth]{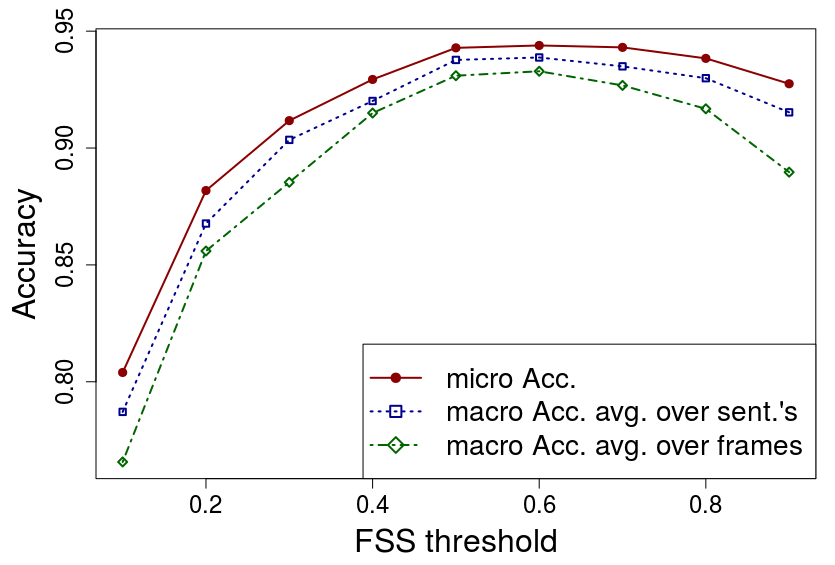}
\end{minipage}
\end{figure*}

\begin{table*}[tb!]
\centering
\caption{Example sentence-word pairs where the top crowd frame choice is different than the expert. The targeted word appears in italics font in the sentence. The frame picked by the expert is marked with $^{(*)}$.}
\label{tab:disagr}

\scalebox{0.85}{
\begin{tabular}{|c|p{14cm}|cc|}
\hline
{\bf \#} & {\bf Sentence} & {\bf Frame} & {\bf FSS} \\ \hline \hline

\multirow{2}{*}{S1} & \multirow{2}{14cm}{Shops {\it aimed} at the tourist market are interspersed with the more workaday ironmongers.} & $aiming$ & 0.808 \\ 
& & $purpose^{(*)}$ & 0.288 \\ \hline

\multirow{2}{*}{S2} & \multirow{2}{14cm}{The major {\it changes} were not to daily tasks and routines , but to the political power base.} & $cause\_change$ & 0.804 \\ 
& & $undergo\_change^{(*)}$ & 0.305 \\ \hline

\multirow{2}{*}{S3} & \multirow{2}{14cm}{This {\it investigation} has been stymied stopped, obstructions thrown up every step of the way.} & $criminal\_investigation$ & 0.898 \\ 
& & $scrutiny^{(*)}$ & 0.377 \\ \hline

\multirow{2}{*}{S4} & \multirow{2}{14cm}{Does supersizing {\it cause} obesity?} & $cause\_to\_start$ & 0.804 \\ 
& & $causation^{(*)}$ & 0.608 \\ \hline

\multirow{2}{*}{S5} & \multirow{2}{14cm}{The loud, raucous Jamaican English dialect and the {\it waving} hands reflect the joy with which social relations are conducted here.} & $body\_movement$ & 0.861 \\ 
& & $gesture^{(*)}$ & 0.463 \\ \hline

\multirow{2}{*}{S6} & \multirow{2}{14cm}{The Intifada {\it heralded} the rise of the Muslim fundamentalism.} & $heralding$ & 0.777 \\ 
& & $omen^{(*)}$ & 0.227 \\ \hline

\multirow{2}{*}{S7} & \multirow{2}{14cm}{Fish (heads discreetly {\it wrapped} in paper) are still hung out to dry in the sun.} & $adorning$ & 0.31 \\ 
& & $filling^{(*)}$ & 0.278 \\ \hline

\end{tabular}
}
\end{table*}

These definitions are mutually dependent, e.g. the definition of SQS depends on the FQS and WQS, the intuition being that low quality workers should not make sentences look bad, and low quality sentences should not make workers look bad, etc.  The mutual dependence requires an iterative  dynamic programming approach, which converged in numerous applications in fewer than 8 iterations.


\section{Crowd vs. Experts}

To evaluate the quality of the crowd annotations, we iterate through different values of thresholds in the FSS to classify a frame-sentence pair as either positive or negative, then compare the results with the annotations of the FrameNet experts. The results for both the micro (i.e. each frame-sentence pair is counted as either true positive, false positive etc. and used in the calculation of the F1 and accuracy) and macro (the F1 and accuracy are calculated for each sentence and each frame, and then averaged into the final values) scores are presented in Figures~\ref{fig:sent_f1} \&~\ref{fig:sent_acc}.


At the best FSS threshold, the accuracy scores are comparable to those presented by \citeauthor{Hong:2011:GCR:2018966.2018970}~\shortcite{Hong:2011:GCR:2018966.2018970}, who report an average accuracy of 0.928, although on a different dataset. However, accuracy in multi-class classification problems are unreliable as there are high numbers of true negatives. The F1 score is likely a more reliable metric of the performance of the crowd, with scores $ > 0.67 $ for all 3 versions of the F1. Finally, an ANOVA test over the paired FSS and expert decision for a frame-sentence pair resulted in the $ F-value = 4597 $ and $p < 2e^{-16} $, proving that there is a statistically significant relationship between the crowd FSS and the decision of the expert.

\begin{table*}[tb!]
\centering
\caption{Different FSS values for the frames $removing$ (P1, P2, P3), $means$ (P4, P5, P6), $attempt\_suasion$ (P7, P8, P9). The targeted word appears in italics font in the sentence. The frame picked by the expert is marked with $^{(*)}$.}
\label{tab:fss_ex}

\scalebox{0.85}{
\begin{tabular}{|c|p{12cm}|c|cc|}
\hline
{\bf \#} & {\bf Sentence} & {\bf SQS} &  {\bf Frame} & {\bf FSS}  \\ \hline \hline

\multirow{3}{*}{P1} & \multirow{3}{12cm}{Egypt has provided no evidence demonstrating the {\it elimination} of its biological warfare ability, which has existed since at least 1972.} & \multirow{3}{*}{0.841} & $removing^{(*)}$ & 0.938 \\
& & & $cause\_change$ & 0.175 \\
& & & $event$ & 0.032 \\ \hline

\multirow{4}{*}{P2} & \multirow{4}{12cm}{First, he forbade seeking the aid of infidels when the Syrian Mujahiddin asked Saddam Hussein to {\it overthrow} the regime of Hafiz Al-Assad in Syria.} & \multirow{4}{*}{0.669} & $change\_of\_leadership^{(*)}$ & 0.847 \\ 
& & & $removing$ & 0.539 \\
& & & $eventive\_cognizer\_affecting$ & 0.087 \\
& & & $people$ & 0.005 \\ \hline

\multirow{5}{*}{P3} & \multirow{5}{12cm}{Their influence helped draw a line in the desert sand between legitimate operations and mob casinos, where illegal {\it skimming} of profits was rampant.} & \multirow{5}{*}{0.366} & $removing^{(*)}$ & 0.532 \\
& & & $theft$ & 0.494 \\
& & & $committing\_crime$ & 0.459 \\
& & & $misdeed$ & 0.431 \\
& & & $cause\_change$ & 0.273 \\ \hline \hline

\multirow{2}{*}{P4} & \multirow{2}{12cm}{The above mentioned protection {\it procedures} are only for observation purposes, while patrols check the fences, the barriers, and the towers.} & \multirow{2}{*}{0.786} & $means^{(*)}$ & 0.889 \\
& & & $being\_employed$ & 0.11 \\ \hline

\multirow{4}{*}{P5} & \multirow{4}{12cm}{We've expanded Goodwill's proven {\it methods} to towns and neighborhoods where they are needed most.} & \multirow{4}{*}{0.364} & $means^{(*)}$ & 0.601 \\
& & & $expertise$ & 0.342 \\
& & & $domain$ & 0.173 \\
& & & $fields$ & 0.131 \\ \hline

\multirow{4}{*}{P6} & \multirow{4}{12cm}{The latest {\it approach} is perhaps the best of the post-mob era : the comprehensive resort.} & \multirow{4}{*}{0.208} & $means^{(*)}$ & 0.457 \\
& & & $conduct$ & 0.225 \\
& & & $path\_traveled$ & 0.159 \\
& & & $communication$ & 0.121 \\ \hline \hline

\multirow{4}{*}{P7} & \multirow{4}{12cm}{Prime Minister Ariel Sharon of Israel {\it urged} President Bush to step up pressure on Iran to give up all elements of its nuclear program.} & \multirow{4}{*}{0.528} & $attempt\_suasion^{(*)}$ & 0.81 \\
& & & $request$ & 0.387 \\
& & & $communication$ & 0.337 \\
& & & $cause\_to\_start$ & 0.115 \\ \hline

\multirow{4}{*}{P8} & \multirow{4}{12cm}{The security team should {\it urge} everyone to take precautions and guard their homes tightly.} & \multirow{4}{*}{0.358} & $attempt\_suasion^{(*)}$ & 0.605 \\
& & & $request$ & 0.321 \\
& & & $cause\_to\_start$ & 0.256 \\
& & & $communication$ & 0.213 \\ \hline

\multirow{3}{*}{P9} & \multirow{3}{12cm}{The security team should publish a periodic bulletin and distribute to all residents, {\it advising} them how to safely store gaz and logs.} & \multirow{3}{*}{0.386} & $attempt\_suasion^{(*)}$ & 0.576 \\
& & & $communication$ & 0.567 \\
& & & $expertise$ & 0.167 \\
& & & $request$ & 0.156 \\ \hline

\end{tabular}
}
\end{table*}

While the majority of expert choices have high FSS scores, there are some exceptions. We observed 3 different causes for this disagreement, which are exemplified in Table~\ref{tab:disagr}:

\begin{enumerate}

\item The crowd {\it misunderstood the frame definition}. For instance, in $S1$, the crowd mistook the $aiming$ frame to mean purpose, instead of the more literal meaning of the frame of adjusting an instrument to reach a target. In $S2$, the crowd correctly identifies a causal sense, but the correct interpretation is a passive change ({\it changes [...] to the political power}) instead of the active change (i.e. a subject is doing the changing) that is picked by the crowd.

\item The {\it information in the sentence is incomplete} to identify the correct frame. $S3$ does not express whether the investigation is criminal in nature, although that is a possible interpretation. This represents a limitation in the design of the crowdsourcing task -- in some versions of the expert task, annotators had the full context of the document available when performing the annotations. This could be fixed or reduced by providing the sentence before and after, without overloading the workers.

\item The crowd offers a {\it legitimate alternative interpretation} of what the correct frame should be. In $S5$ the crowd picks the more general frame $body\_movement$ for {\it waving}, while in $S4$ and $S6$, the crowd picks more specific interpretations than the expert ($cause\_to\_start$ for the {\it obesity} effect instead of the broader sense of $causation$ in $S4$, and $heralding$ instead of $omen$ for the word {\it heralding} in $S6$). $S7$ shows an example where the expert made a mistake, as $filling$ refers to the action of covering an area with something, whereas $adorning$ refers to the passive act of being covered.

\end{enumerate}

\section{Capturing Ambiguity}

The cases where the experts and crowd disagree exemplify how difficult frame disambiguation can be when dealing with ambiguity, both in sentences and in the frame definition. Currently in the FrameNet corpus, the expert annotations lack the level of granularity necessary to differentiate between clear expressions of the frames, and more ambiguous ones. We propose the FSS metric as a method to capture the degree of ambiguity with which a frame captures a word sense in a sentence. In Table~\ref{tab:fss_ex}, we show how the FSS metric varies together with the clarity with which a frame is expressed across different sentences. We demonstrate this across 3 different frames:

\begin{itemize}

\item $removing$: $P1$ is an unambiguous expression of the frame, as reflected by the high agreement score. In $P2$, the top crowd frame as well as the expert choice frame $change\_of\_leadership$ refers to overthrowing the government, and $removing$ can be read as a generalization of this sense (i.e. removing the government by overthrowing it) -- $removing$ is a valid interpretation, but less specific, and the lower FSS seems justified. $P3$ is an even more ambiguous case -- it is not clear whether the word {\it skimming} refers to generally $committing\_crime$, or to the more specific crime of $theft$, and $removing$ is a generalization for the sense of $theft$, however skimming here is a common metaphor, and not the actual act of skimming. We claim the rank ordering of uses of the {\it removing} frame here is sensible, moreover it is far more useful to capture this information than require a single discrete truth value - the third case is simply not as clear a usage of the frame as the first.  There is a certain arbitrariness to determining which of these is "truly removing" and which is not.

\begin{figure*}[tbh!]
\centering
\begin{minipage}{.471\textwidth}
\centering
\caption{SQS in relation to F1 score (with expert annotations as true positives), shows that in higher quality sentences, the crowd tends to agree with experts.}
\label{fig:sqs_f1}
\includegraphics[width=\textwidth]{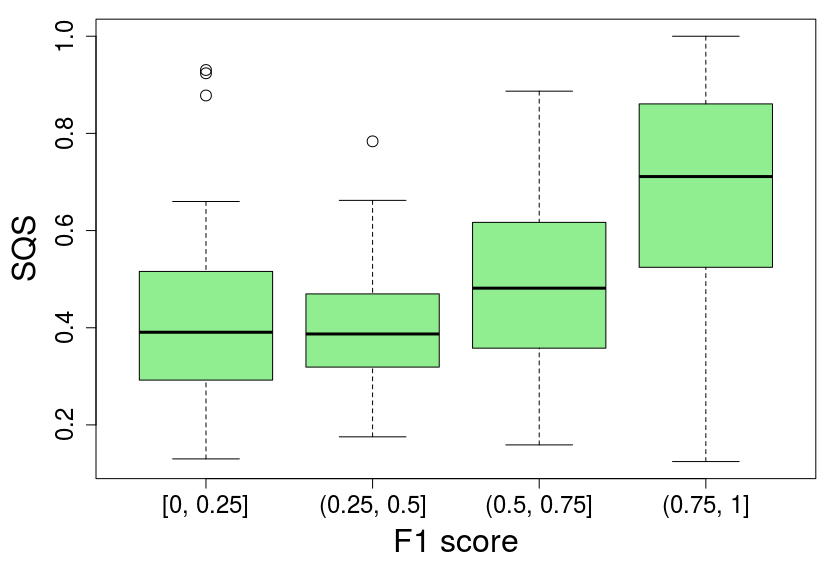}
\end{minipage} \qquad 
\begin{minipage}{.471\textwidth}
\centering
\caption{FQS in relation with F1 score, shows that in higher quality frames, the crowd tends to agree with the experts.}
\label{fig:fqs_f1}
\includegraphics[width=\textwidth]{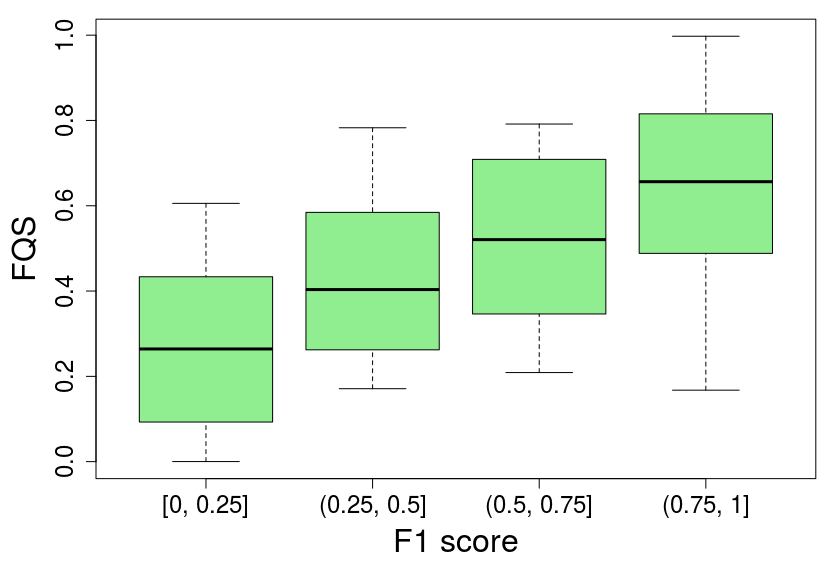}
\end{minipage}
\end{figure*}

\begin{table*}[tbh!]
\centering
\caption{Sentence Quality Score Examples. The targeted word appears in italics font in the sentence. The frame picked by the expert is marked with $^{(*)}$.}
\label{tab:sqs}

\scalebox{0.85}{
\begin{tabular}{|c|p{12cm}|c|cc|}
\hline
{\bf \#} & {\bf Sentence} & {\bf SQS} &  {\bf Frame} & {\bf FSS}  \\ \hline \hline

\multirow{3}{*}{Q1} & \multirow{3}{12cm}{Although David bought the land for the Temple and carefully assembled its building materials, he was deemed unworthy of {\it constructing} the Temple.} & \multirow{3}{*}{0.711} & $building^{(*)}$ & 0.925 \\
& & & $manufacturing$ & 0.183 \\ 
& & & $create\_physical\_artwork$ & 0.056 \\ \hline

\multirow{3}{*}{Q2} & \multirow{2}{12cm}{Passageways for cars and pedestrians should be designated 4- Road bumps: Six successive bumps should be {\it constructed} at 500 meters from the location.} & \multirow{3}{*}{0.542} & $building^{(*)}$ & 0.768 \\
& & & $manufacturing$ & 0.326 \\ 
& & & $create\_physical\_artwork$ & 0.089 \\ \hline

\multirow{3}{*}{Q3} & \multirow{3}{12cm}{{\it Constructed} in wood, brick, stone, ceramic, and bronze, this is a work of extravagant beauty, uniting many ancient art forms.} & \multirow{5}{*}{0.351} & $building^{(*)}$ & 0.515 \\
& & & $create\_physical\_artwork$ & 0.335 \\
& & & $manufacturing$ & 0.237 \\ \hline \hline

\multirow{3}{*}{Q4} & \multirow{3}{12cm}{U.S. Congressman Tony Hall arrived here Sunday evening, {\it becoming} the first U.S. lawmaker to visit Iraq since the 1991 Gulf War.} & \multirow{3}{*}{0.901} & $becoming^{(*)}$ & 0.995 \\
& & & $cause\_change$ & 0.24 \\
& & & $undergo\_change$ & 0.212 \\ \hline

\multirow{3}{*}{Q5} & \multirow{3}{12cm}{Cheung Chau {\it becomes} the center of Hong Kong life once a year, usually in May , during the Bun Festival, a folklore extravaganza.} & \multirow{3}{*}{0.562} & $becoming^{(*)}$ & 0.783 \\
& & & $undergo\_change$ & 0.783 \\ 
& & & $cause\_change$ & 0.402 \\ \hline \hline

\multirow{2}{*}{Q6} & \multirow{3}{12cm}{Are there any {\it efforts} to bring back small investors?} & \multirow{2}{*}{0.811} & $attempt^{(*)}$ & 0.926 \\
& & & $commitment$ & 0.178 \\ \hline

\multirow{2}{*}{Q7} & \multirow{2}{12cm}{At AOL there was a conscious {\it effort} to develop other ``characters,'' for lack of a better word.} & \multirow{2}{*}{0.588} &  $attempt^{(*)}$ & 0.739 \\
& & & $commitment$ & 0.468 \\ \hline

\end{tabular}
}
\end{table*}

\item $means$: This frame refers to the means used by an agent to achieve a purpose. While $P4$ offers an unambiguous expression of the frame, in $P5$ the means with which to achieve a goal becomes confused with the expertise and knowledge required to achieve it. In $P6$ the goal is not mentioned, therefore creating confusion about the purpose of the {\it approach}, and whether it might refer to a way of communicating or behaving.  Again, we claim this rank ordering is more informative than requiring a discrete judgment on each case.

\item $attempt\_suasion$: This frame refers to a speaker attempting to influence the addressee to act. Sentences $P7$ to $P9$ express various degrees of persuasion, from obviously to weakly expressed. In $P7$, it is clear that the attempt at persuasion is an event that has occurred ({\it Sharon [...] urged}). $P8$ expresses an obligation at an attempt to persuade ({\it should urge}), whereas in $P9$ the persuasion is weaker, merely {\it advice}.

\end{itemize}

In addition to the ranking, the method of collecting data from multiple crowd workers yields alternate interpretations, which are also quite useful.  Consider that a common motivation for collecting annotated data is to train and evaluate deep learning models, many of which produce vectors of output (frame disambiguation can be implemented as a multi-class problem).  Our methods of gathering annotations are naturally suited to multi-class objectives.

The SQS and FQS metrics can additionally be used to express the overall ambiguity in the sentence and frame, respectively. Figures~\ref{fig:sqs_f1} \&~\ref{fig:fqs_f1} show that sentences with higher SQS and frames with higher FQS also have higher F1 values, demonstrating that the SQS and FQS metrics can be useful in determining data quality. This result, in combination with the correlation between FSS and expert annotations, shows that when there is agreement in the crowd, then the crowd also agrees with the experts, but when there is disagreement, it may be because something is wrong: with the workers, the sentence, or the frames.

In Table~\ref{tab:sqs}, we show some examples of how SQS captures the clarity for the sense of a word in a sentence, by taking the same word (and therefore same list of candidate frames) in different sentences:

\begin{table*}[tb!]
\centering
\caption{Frame Quality Score Examples. The targeted word appears in italics font in the sentence.}
\label{tab:fqs}

\scalebox{0.85}{
\begin{tabular}{|c|c|p{5cm}|p{10cm}|c|}
\hline
{\bf Frame} & {\bf FQS} & {\bf Definition} & {\bf Example Sentences} & {\bf FSS} \\ \hline \hline

\multirow{2}{*}{$killing$} & \multirow{2}{*}{0.954} & \multirow{2}{5cm}{A Killer or Cause causes the death of the Victim.} &  $F1$: Older kids left homeless after a recent murder-{\it suicide} in Indianapolis claimed Mom and Dad. & 0.8 \\
& & & $F2$: The incident at Mayak was the third {\it shooting} in recent weeks involving nuclear weapons or facilities in Russia. & 0.75 \\ \hline

\multirow{3}{*}{ $food$ } & \multirow{3}{*}{0.838} & \multirow{3}{5cm}{Words referring to items of food.} & $F3$: Lamma Island is perfect for sitting back to watch {\it bananas} grow. & 1.0 \\
& & & $F4$: Along with the usual {\it chickens}, you will see for sale snakes, dogs, and sometimes monkeys - all highly prized delicacies . & 0.838 \\
& & & $F5$: You can browse among antiques, flowers, {\it herbs}, and more. & 0.503 \\ \hline

\multirow{3}{*}{ $assistance$ } & \multirow{3}{*}{ 0.634 } & \multirow{3}{5cm}{A Helper benefits a Benefited party by enabling the culmination of a Goal of the Benefited party.} & $F6$: Your support {\it helps} provide real solutions. & 0.955 \\ 
& & & $F7$: Unemployment {\it provides} benefits that many entry-level jobs don't. & 0.467 \\ 
& & & $F8$: Your support of Goodwill will {\it provide} job training. & 0.401 \\ \hline

\multirow{3}{*}{ $purpose$ } & \multirow{3}{*}{ 0.63 } & \multirow{3}{5cm}{An Agent wants to achieve a Goal. A Means is used to allow the Agent to achieve a Goal.} & $F9$: The {\it objective} of having kiosks is they serve as communication points between the guards & 0.94 \\
& & & $F10$: They are antiviral drugs {\it designed} to shorten the flu. & 0.476 \\ 
& & & $F11$: It seems that the city produced artists of this stature by accident, even against its {\it will}. & 0.241 \\ \hline

$subjective$ & \multirow{3}{*}{0.366} & \multirow{3}{5cm}{An Agent has influence on a Cognizer. The influence may be general, manifested in an Action as a consequence of the influence.} & $F12$: There have been changes, many of them due to economic progress, new construction, and other factors that {\it influence} cities. & 0.54 \\
$influence$ & & & $F13$: The Cycladic culture was {\it influenced} by societies in the east. & 0.46 \\
& & & $F14$: Their complaint: the system {\it discourages} working. & 0.364 \\ \hline

$undergo$ & \multirow{3}{*}{0.313} & \multirow{3}{5cm}{An Entity changes, either in its category membership or in terms of the value of an Attribute.} & $F15$: The animosity between these two traditional enemies is beginning to {\it diminish}. & 0.805 \\ 
 $change$ & & & $F16$: The {\it shift} in the image of Gates has been an interesting one for me to watch. & 0.351 \\
& & & $F17$: The settlements of Thira and Akrotiri {\it thrived} at this time. & 0.256 \\ \hline

\end{tabular}
}
\end{table*}

\begin{itemize}

\item Sentences $Q1$, $Q2$ and $Q3$ all contain the word {\it construct}, with different degrees of clarity. When the object being constructed is a building (i.e. the {\it Temple} in $Q1$), there is no ambiguity in selecting the $building$ frame, but when the object is a {\it road bump} ($Q2$), the sense of the building $frame$ becomes difficult to separate from $manufacturing$. In $Q3$, the object of the construction is not expressed, but the construction materials imply a precious object, therefore $building$, $manufacturing$ and $create\_physical\_artwork$ are all possible interpretations. Sentences 

\item $Q4$ and $Q5$ illustrate the variation in clarity for the word {\it become}. While in $Q4$, the sense $becoming$ is the unambiguous choice, in $Q5$ it is difficult to choose between the frames $becoming$ and $undergo\_change$ (it is arguable that {\it Cheung Chau} needs to undergo some form of change in order to become a center).

\item $Q6$ and $Q7$ both deal with the word {\it effort}. In $Q7$, however, the {\it conscious} qualifier for the word {\it effort}, as well as the goal to {\it develop}, implies a sustained, long-term action that can be understood as either an $attempt$ or a $commitment$ to achieve a goal. In contrast, $Q6$ expresses a short-term, concrete action (to {\it bring}), which more closely fits the sense of the frame $attempt$.

\end{itemize}

Again, our claim is that these scores and ranking are far more sensible and informative than requiring a discrete truth decision, which seems more arbitrary as the scores decrease.

As the examples above indicate, one possible cause for sentence ambiguity is missing context information (e.g. in $Q3$). This was also one of the causes for disagreement between crowd and expert. A solution to this problem would be to expand the input text for the crowdsourcing task, to include the full paragraph, or even just one sentence before and one after the one we want the crowd to annotate.

Another reason for sentence ambiguity is frames that overlap in meaning (e.g. in $Q5$ and $Q7$). While providing more context could help with this, it is often the case that even the definitions of the frames are very close. The FQS metric is a useful indicator for these case.

Table~\ref{tab:fqs} shows varying FQS values for different frames, from very clear to ambiguous. The frame $subjective\_influence$, with an FQS of 0.366, has a low score compared to the others. From looking at the sentences, we observed that the crowd had difficulty distinguishing between this frame and $objective\_influence$. The difference between these two frames is very small -- $subjective\_influence$ means a general, vague type of influence, whose effect cannot be measured, whereas $objective\_influence$ refers to a more concrete type of influence. However, as we see from the example sentences in Table~\ref{tab:fqs}, these cases can be very difficult to separate in natural language (e.g. in $F13$ is {\it cultural influence} subjective or objective?).

Another feature we observed was the correlation of FQS with how abstract the sense of the frame is. Frames with high FQS, such as $killing$ and $food$, tend to refer to concrete events or objects. These frames can still appear in ambiguous contexts (e.g. in $F5$, it is not clear whether {\it herbs} classify as a type of $food$), but overall these frames refer to specific and particular senses that are unambiguous. As the value of the FQS metric goes down, the frames become more abstract. $assistance$ and $purpose$ both have example sentences where they are expressed unambiguously ($F6$ and $F9$), but their definitions are more abstract, and therefore have more room for interpretation. For instance, providing benefits (in $F7$) or expertise (in $F8$) can be regarded as a type of help, or $assistance$, even though the expert picked the more literal sense of the frame $supply$ for both of these cases. Likewise the frame $purpose$ can be understood in $F10$ as the purpose of a design (the expert picked the more literal $coming\_up\_with$), or in $F11$ as the goal of the desire/will (the expert picked $desiring$). $undergo\_change$, the frame with the lowest FQS in Table~\ref{tab:fqs} has a very broad meaning, and is a generalization of other more specific frames: $change\_position\_on\_a\_scale$ in $F16$, and $thriving$ in $F17$.

As we have seen from these examples, ambiguity in frames is connected to ambiguity in sentences. Frames with abstract or overlapping definitions are likely to appear in ambiguous sentences, and missing context from sentences is likely to result in more ambiguous scores for the frames. While workers misunderstanding the task is also a confounding factor that adds to the noise in the data, it is clear that there are many instances where inter-annotator disagreement is legitimately a by-product of ambiguity. This is an issue with the FrameNet dataset, as it does not allow for expressing the various degrees with which a sense applies to a word in a sentence, and instead relies on binary labels (i.e. the frame is expressed or not). This results in a loss of information that could impact the various natural language processing and machine learning applications that make use of this corpus, as it sets false targets for optimization -- i.e. it seems unfair to expect a model to differentiate between highly ambiguous examples, when even human annotators are having such difficulty with them.

\section{Conclusion}

In this paper, we present an approach to crowdsource frame disambiguation annotations in sentences. We adapted an existing method, CrowdTruth~\cite{aroyo2014threesides}, that uses multiple workers per sentence, in order to capture and interpret inter-annotator \emph{disagreement} as an indication of ambiguity. We modified CrowdTruth metrics in order to capture frame-sentence agreement (FSS), sentence quality (SQS) and frame quality (FQS). We performed an experiment over a set of 433 sentences annotated with frames from FrameNet corpus, and showed that the aggregated crowd annotations achieve an F1 score greater than 0.67 compared to expert linguists, and an accuracy that is comparable to the state of the art~\cite{Hong:2011:GCR:2018966.2018970}. It is our intention to scale out the task on new data to provide an ambiguity-enhanced dataset for experimentation.

We showed cases where the crowd annotation is correct even though the expert is in disagreement, arguing for the need to have multiple annotators per sentence. Most importantly, we examined the cases in which crowd workers could not agree. We found that disagreement is caused by one or more of the following: workers misunderstanding the task, missing context from the sentences, frames with overlapping or abstract definitions. The results show a clear link between inter-annotator disagreement and ambiguity, either in the sentence, frame, or the task itself. We argue that collapsing such cases to a single, discrete truth value (i.e. correct or incorrect) is inappropriate, creating brittle, incomplete datasets, and therefore arbitrary targets for machine learning.  We further argued that ranking examples by a score is informative, and that the crowd offers alternate interpretations that are often sensible.

\section*{Acknowledgments}

We would like to thank Luigi Asprino, Valentina Presutti and Aldo Gangemi for their assistance with using the Framester corpus, as well as their advice in better understanding the task of frame disambiguation. We would also like to thank the anonymous crowd workers for their contributions to our crowdsourcing tasks.

\bibliographystyle{aaai}
\bibliography{sources}

\end{document}